\documentclass[letterpaper, 10 pt, conference]{ieeeconf}  

\IEEEoverridecommandlockouts                              
\usepackage {graphicx}                                                          
\usepackage{amssymb} 
\usepackage{xcolor}  
\usepackage{pifont}  
\usepackage{multirow}
\usepackage{adjustbox}
\usepackage{booktabs}
\usepackage{amsmath}
\usepackage {float} 
\usepackage{dblfloatfix}
\usepackage{graphicx}
\usepackage{caption}
\usepackage{booktabs}
\usepackage[top=2cm, bottom=2.05cm, left=1.92cm, right=1.92cm]{geometry}

\title{\LARGE \bf
DVDP: An End-to-End Policy for Mobile Robot Visual Docking with RGB-D Perception
}
\begin{document}
\author{%
    Haohan Min\textsuperscript{\textdagger}, 
    Zhoujian Li\textsuperscript{\textdagger}, 
    Yu Yang, 
    Jinyu Chen, 
    Shenghai Yuan\textsuperscript{*}%
    \thanks{\textsuperscript{1}These authors contributed equally to this work.}
    \thanks{\textsuperscript{*}Corresponding author.}
}
\twocolumn[{
\renewcommand\twocolumn[1][]{#1}

\maketitle
\vspace{-10mm} 

\begin{center}
     \captionsetup{type=figure}
    \includegraphics[width=0.93\textwidth]{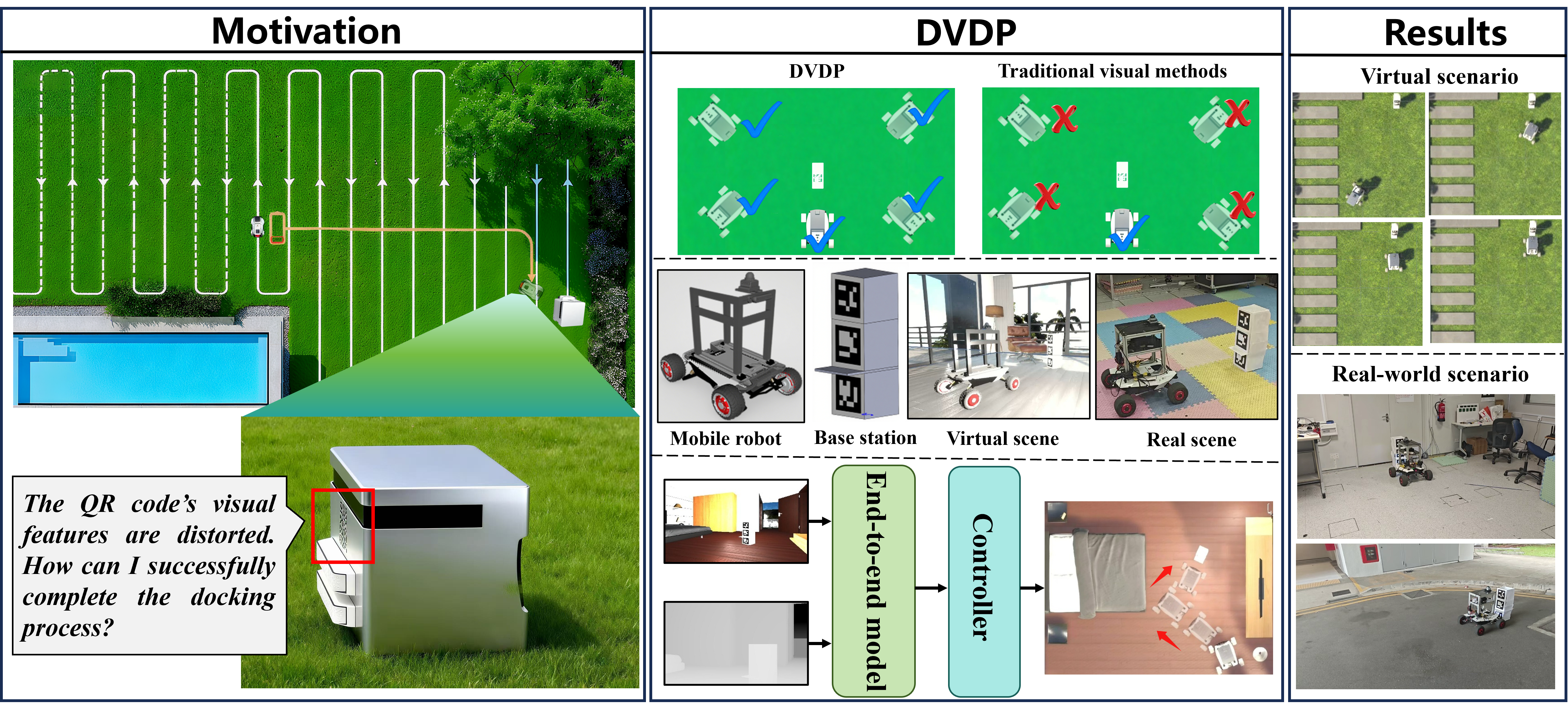}
    \caption{Overview of our paper. The first column presents our motivation, the second column details the design of the end-to-end visual docking strategy, and the third column showcases selected experimental results, including both virtual and real-world scenarios.}
\end{center}
}]

{ 

\makeatletter 
\def\thefootnote{} 
\footnotetext{
\textsuperscript{\textdagger}These authors contributed equally to this work.

\textsuperscript{*}Corresponding author. E-mail: shyuan@ntu.edu.sg

Haohan Min is with the Shenzhen International Graduate School, Tsinghua University. Zhoujian Li is with the College of Design and Engineering, National University of Singapore. Yu Yang, Jinyu Chen, and Shenghai Yuan are with the School of Electrical and Electronic Engineering, Nanyang Technological University.
}
\makeatother 
} 

\thispagestyle{empty}
\pagestyle{empty}

\begin{abstract}Automatic docking has long been a significant challenge in the field of mobile robotics. Compared to other automatic docking methods, visual docking methods offer higher precision and lower deployment costs, making them an efficient and promising choice for this task. However, visual docking methods impose strict requirements on the robot's initial position at the start of the docking process. To overcome the limitations of current vision-based methods, we propose an innovative end-to-end visual docking method named DVDP(direct visual docking policy). This approach requires only a binocular RGB-D camera installed on the mobile robot to directly output the robot's docking path, achieving end-to-end automatic docking. Furthermore, we have collected a large-scale dataset of mobile robot visual automatic docking dataset through a combination of virtual and real environments using the Unity 3D platform and actual mobile robot setups. We developed a series of evaluation metrics to quantify the performance of the end-to-end visual docking method. Extensive experiments, including benchmarks against leading perception backbones adapted into our framework, demonstrate that our method achieves superior performance. Finally, real-world deployment on the SCOUT Mini confirmed DVDP's efficacy, with our model generating smooth, feasible docking trajectories that meet physical constraints and reach the target pose.

\end{abstract}

\section{INTRODUCTION}

Autonomous docking is essential for long-term reliable operation of mobile robots, enabling automatic charging and data synchronization without human intervention. It underpins persistent autonomy across domains such as logistics, agriculture, and domestic services, where continuous deployment is required. As mobile robots proliferate, robust docking has become a cornerstone technology for scalable and practical applications.

\textbf{Existing} automated docking policies include those based on infrared sensors \cite{zhou2023autonomous}, ultrasonic sensors \cite{2015Calibration}, and visual sensors \cite{2024A,2025Machine}. Among these, vision-based methods have emerged as particularly promising due to their ability to provide rich environmental information and achieve precise localization through image processing and computer vision techniques \cite{2024Vision}. They are also relatively easy to implement and deploy, making them attractive for a wide range of robotic platforms \cite{2024Vision11}. However, most vision-based docking methods adopt a perception–planning–control pipeline: detecting feature points on the docking station, estimating robot pose, and subsequently generating motion commands \cite{Zhang:22}. Such pipelines are inherently sensitive to errors in each stage, especially when the robot’s initial state deviates significantly from the expected docking line \cite{2022Automated,2019Autonomous}.

The key \textbf{challenge} is that mobile robots often return to the docking station after extended tasks, during which localization may drift or positions deviate. This is especially critical outdoors, such as golf courses or agricultural fields, where uneven terrain, GPS errors, and odometry drift cause large deviations from the docking path. In such cases, visual features are easily distorted or occluded, leading to inaccurate perception. These errors propagate to planning and control, often causing docking failures. Existing approaches therefore, lack robustness to arbitrary initial poses and outdoor uncertainties, fundamentally limiting their reliability.

To address these limitations, we propose an end-to-end mobile robot docking policy named \textbf{DVDP}. As illustrated in Fig.~1, our framework takes synchronized RGB and depth images from an onboard camera as input and directly outputs the docking trajectory, eliminating the need for handcrafted feature extraction or rule-based intermediate modules. This design enables robust docking from arbitrary initial positions, overcoming the strict constraints of traditional perception–planning–control methods. Furthermore, we construct a hybrid indoor–outdoor docking dataset to support future research in this domain. To the best of our knowledge, this is the first end-to-end, model-driven approach to vision-based visual docking for mobile robots, validated through both large-scale datasets and real-robot deployment.

\noindent\textbf{Our contributions are as follows:}
\begin{itemize}
    \item We introduce DVDP, a novel end-to-end model-driven policy for visual docking in mobile robots. DVDP effectively addresses the limitations of existing visual docking algorithms that require strict initial positions at the start of docking, enabling reliable task completion from any initial position.
    
    \item We have constructed a large-scale dataset for DVDP and future end-to-end visual docking strategies. This dataset includes a wide range of indoor and outdoor hybrid scenarios. 
    
    \item  Extensive comparative and ablation experiments demonstrate that DVDP achieves the best performance in current end-to-end visual docking tasks. Furthermore, deployment on the SCOUT Mini platform validated DVDP's effectiveness, demonstrating that our policy produces smooth, feasible trajectories within constraints that accurately achieve the target pose.
\end{itemize}

\begin{figure*}[htbp] 
    \centering 
    \includegraphics[width=0.93\textwidth]{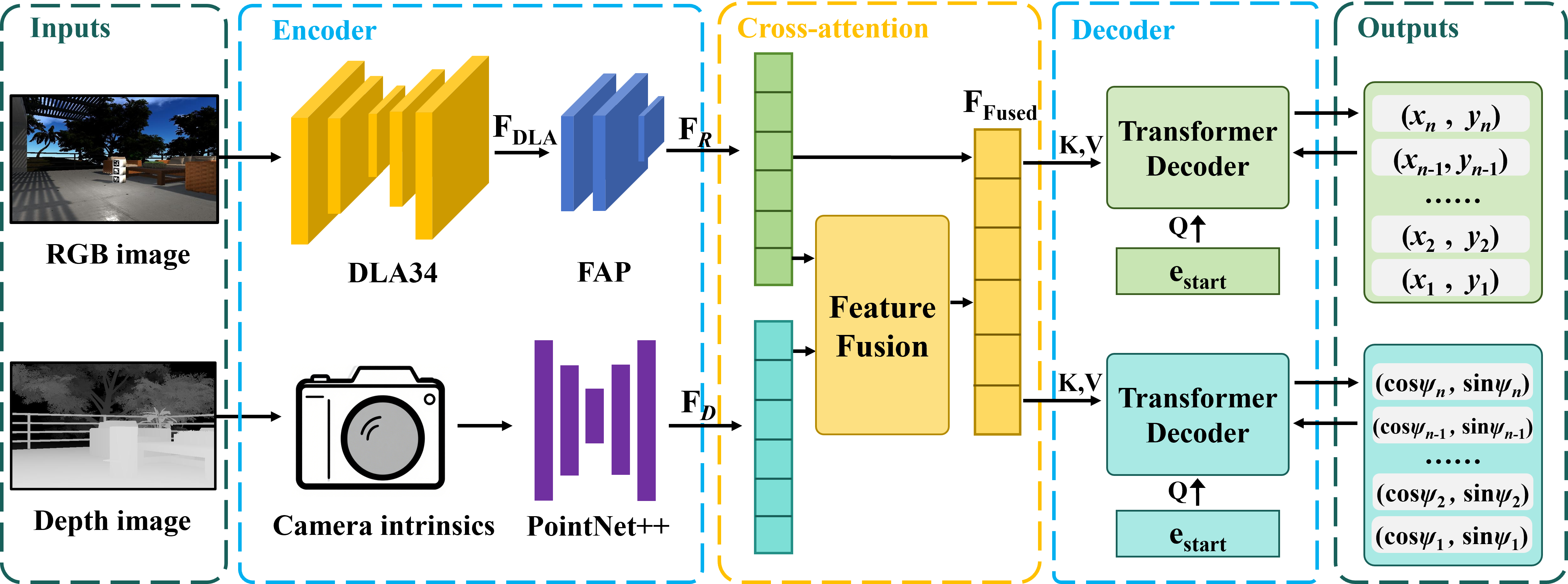} 
    \caption{Architecture of DVDP network. An RGB stream (DLA+FAP) extracts pose information, while a depth stream (PointNet++) captures geometric context. These features are fused via cross-attention and then fed into two decoders to output the final trajectory points and orientation sequence.}
    \label{fig:two_column_center}
    \vspace{-5mm} 
\end{figure*}
\setlength{\textfloatsep}{5pt} 

\section{RELATED WORKS}

\subsection{Vision-Based Automatic Docking Methods}
Current vision-based auto-docking methods rely on recognizing and detecting feature points on the base station, such as shapes, colors, markers, or textures. By utilizing these features, the system can estimate the robot's pose relative to the base station. This information is then used for path planning, allowing the robot to navigate safely and accurately to the base station while considering environmental obstacles, thereby ensuring a reliable automatic docking function. \cite{2022Automated} develops a vision-guided system for UAV aerial docking using a lightweight perception approach. Their method employs YOLOv4-tiny and an RGB-D camera to estimate the target's 3D position in real-time. However, this approach suffers from reduced accuracy under large viewing angles. \cite{Zhang:22} proposes a vector heuristic visual guidance method. This approach leverages the scattering effects of four laser beams in water to construct virtual "wing light" markers. It combines the Hough transform with the P3P algorithm to solve the six degrees of freedom pose estimation. However, the method's strong reliance on water scattering limits its applicability to underwater environments, thus reducing its generalizability. \cite{Volden2022Visionbased} introduces a hybrid method for USV autonomous docking, using YOLO for robust target detection and the ArUco algorithm for precise 3D pose estimation. However, the method's primary limitation is its reliance on pre-deployed markers, which restricts its use in marker-less environments. \cite{2019Autonomous} designs a vision-guided docking system for AUVs, utilizing an L-shaped LED light array as a visual beacon. The system uses image processing to determine real-time position and heading for docking. However, its accuracy significantly diminishes with large AUV roll angles, affecting pose estimation and applicability in complex conditions. Overall, most vision-based docking algorithms use a perception-planning-control framework, relying on visual markers on the base station. This dependence often requires strict robot initial poses to prevent visual data distortions that could impact performance and accuracy.

\subsection{End-to-end Model Driven Robot Tasks}
In recent years, the application of end-to-end models in the field of intelligent robotics has made significant progress. Broadly speaking, these models are primarily utilized in end-to-end navigation, end-to-end grasping, and other related areas. Below, I will review several research works in each of these domains. 

End-to-end navigation: \cite{perez2021robot} explores the use of LiDAR-based mapping as input for training end-to-end deep reinforcement learning (DRL) networks. This network aims to achieve precise local navigation and dynamic obstacle avoidance.  \cite{kulhanek2021visual} develops a vision-based navigation method. Through extensive training using DRL in a simulated environment, the well-trained policy can be effectively deployed on real-world robots. ENTL \cite{Kotar_2023_ICCV} integrates world modeling, localization, and imitation learning into a unified vision-based sequential prediction task. This approach facilitates long-sequence representation for embodied navigation. \cite{liu2024volumetric} combines language prompts with visual inputs to enhance the performance of end-to-end navigation in robotics.

End-to-end grasping: LEGATO \cite{10855557} transfers grasping skills across robots with different morphologies. This approach uses a hand-held gripper to learn a grasping policy in a kinematic-invariant space, separating the skill from the demonstrator's body. \cite{10854684} addresses poor generalization of robotic grasping policies by introducing a diffusion-based method. This method allows cross-gripper transfer by encoding the scene in a gripper-agnostic way and decoding grasp poses tailored to the gripper's geometry. \cite{liu2024efficient} presents an end-to-end 6-DoF bin picking method that learns a diverse grasp distribution, using a power sphere representation trained on real samples, ensuring robustness to noise in depth images.

Other related areas: FusedNet \cite{Chen2024FusedNet}, an end-to-end network for mobile robot relocalization in dynamic environments, uses a monocular camera and cross-attention to integrate global and local features, improving 3-DoF pose estimation accuracy.  DCT \cite{PAL2024104567}, a vision-based end-to-end architecture for detection, tracking, and classification, enables robots to assist human pickers in fruit harvesting tasks. Extreme Parkour \cite{10610200} utilizes visual input from a single depth camera and employs an end-to-end reinforcement learning framework for a low-cost quadruped robot. This system translates visual data into motor commands for executing complex tasks, enabling the robot to perform extreme parkour maneuvers.

\section{METHODOLOGY}
\subsection{Problem Formulation}
\label{sec:problem_formulation}

Our goal is to train an end-to-end deep neural network, denoted as $\mathcal{N}_{\theta}$, which takes the RGB image $I$ and the corresponding depth image $P$ as inputs, and predicts the mobile robot's future position and orientation trajectories.

Formally, our dataset $\mathcal{D}$ is defined as a collection of samples, where each sample consists of an RGB image, a depth image, and the corresponding ground truth position and orientation sequences:
\begin{equation}
    \mathcal{D} = \{ (I_k, P_k, \mathcal{P}_k^{\text{gt}}, \mathcal{O}_k^{\text{gt}}) \}_{k=1}^M
    \label{eq:dataset}
\end{equation}
where $k$ is the sample index and $M$ is the total number of samples in the dataset.
\begin{itemize}
    \item $I_k \in \mathbb{R}^{H \times W \times 3}$ is an RGB image.
    \item $P_k \in \mathbb{R}^{H \times W}$ is a single-channel depth image.
    \item $\mathcal{P}_k^{\text{gt}}$ represents the ground truth future position trajectory, which consists of $Q$ 2D coordinate points:
    \begin{equation}
        \mathcal{P}_k^{\text{gt}} = \{ p_j^{\text{gt}} \}_{j=1}^Q = \{ (X_j, Y_j) \}_{j=1}^Q
        \label{eq:gt_position_trajectory}
    \end{equation}
    \item $\mathcal{O}_k^{\text{gt}}$ represents the corresponding future orientation trajectory, which consists of $Q$ unit direction vectors:
    \begin{equation}
    \begin{gathered}
        \mathcal{O}_k^{\text{gt}} = \{ o_j^{\text{gt}} \}_{j=1}^Q = \{ (\cos\psi_j, \sin\psi_j) \}_{j=1}^Q \\
        \psi_j \in [-\pi, \pi]
    \end{gathered}
    \label{eq:gt_orientation_trajectory}
\end{equation}
\end{itemize}

The network $\mathcal{N}_{\theta}$ aims to learn a mapping function that concurrently predicts the position and orientation sequences from the sensory inputs:
\begin{equation}
    (\mathcal{P}_k^{\text{pred}}, \mathcal{O}_k^{\text{pred}}) = \mathcal{N}_{\theta}(I_k, P_k)
    \label{eq:network_mapping}
\end{equation}
where the predicted position sequence $\mathcal{P}_k^{\text{pred}}$ and orientation sequence $\mathcal{O}_k^{\text{pred}}$ are structurally identical to their ground truth counterparts.

\subsection{End-to-end visual docking network}

1) Overview: As depicted in Fig. 2, the DVDP network features a dual-encoder architecture to process RGB-D input. For the RGB modality, a DLA backbone \cite{yu2018deep} combined with a \emph{Feature Aggregation Pipeline}(FAP) extracts appearance-based features crucial for identifying the base station. Concurrently, a PointNet++ encoder processes the depth image to capture precise geometric information for localization and environmental context for collision avoidance. These multi-modal features are then integrated using a cross-attention mechanism. Finally, the fused representation is fed into two separate decoders that generate the planned trajectory points and the corresponding robot orientations.

\textbf{Encoder:} Let the input RGB image be denoted as 
\(\mathbf{I} \in \mathbb{R}^{H \times W \times 3}\), 
which is resized to a canonical resolution of 
\(H = W = 512\) to conform with the network’s requirements. 
We adopt DLA-34 as the backbone \cite{9812299}, owing to its 
hierarchical feature aggregation strategy that effectively fuses 
multi-level representations. After processing through DLA, the 
feature tensor becomes 
\(\mathbf{F}_{\text{DLA}} \in \mathbb{R}^{128 \times 128 \times 64}\). 

To distill a high-dimensional, context-rich embedding, we introduce 
a \emph{Feature Aggregation Pipeline (FAP)}. At its core lies the 
Pyramid Pooling Module (PPM) \cite{zhao2017pyramid}, which pools 
\(\mathbf{F}_{\text{DLA}}\) at multiple spatial granularities 
(\(1 \times 1, 2 \times 2, 3 \times 3, 6 \times 6\)), yielding a 
multi-scale representation that encodes both global context and 
local details. This multi-scale tensor is concatenated and 
subsequently condensed via a Global Average Pooling (GAP) operation, 
producing a compact descriptor 
\(\mathbf{f}_{\text{GAP}} \in \mathbb{R}^{d}\), 
with \(d = 2048\). Finally, an MLP is employed to learn complex 
non-linear feature interactions, projecting \(\mathbf{f}_{\text{GAP}}\) 
into the final embedding space:
\begin{align}
\mathbf{z} = \text{MLP}(\mathbf{f}_{\text{GAP}}), \quad 
\mathbf{z} \in \mathbb{R}^{2048}.
\end{align}


\textbf{Depth Stream:}
Let the raw depth map be \(\mathbf{P}\in\mathbb{R}^{H\times W}\) and its vectorized form
\(\mathbf{p}=\operatorname{vec}(\mathbf{P})\in\mathbb{R}^{HW}\).
Define the homogeneous pixel grid
\(\tilde{\mathbf{U}}\in\mathbb{R}^{3\times HW}\) with columns
\(\tilde{\mathbf{u}}_i=[u_i,\,v_i,\,1]^\top\).
Given camera intrinsics \(\mathbf{K}\), we precompute and \emph{cache} the
\emph{back-projection direction matrix}
\begin{align}
\mathbf{D}\;\triangleq\;\mathbf{K}^{-1}\tilde{\mathbf{U}}\;\in\;\mathbb{R}^{3\times HW},
\end{align}
which is re-used across iterations to avoid repeated inversions.
We adopt a uniform sub-sampling operator
\(\mathbf{S}\in\{0,1\}^{N\times HW}\) (e.g., stride \(s=5\) along each axis),
that selects \(N\) pixels indexed by \(\Omega\), yielding
\(\mathbf{p}_\Omega=\mathbf{S}\mathbf{p}\in\mathbb{R}^{N}\) and
\(\mathbf{D}_\Omega=\mathbf{D}\mathbf{S}^\top\in\mathbb{R}^{3\times N}\).
Let the metric depth vector be \(\mathbf{z}=\mathbf{p}_\Omega/s_{\text{depth}}\). The \emph{cached back-projection} into the camera frame is then
\begin{equation}
\label{eq:cached_backproj}
\mathbf{X}\;=\;\mathbf{D}_\Omega\,\operatorname{Diag}(\mathbf{z})
\;\in\;\mathbb{R}^{3\times N},
\end{equation}
whose columns are \(\mathbf{x}_i=z_i\,\mathbf{K}^{-1}\tilde{\mathbf{u}}_i\).
When an extrinsic transform \(\mathbf{T}_{CB}=[\mathbf{R}\,|\,\mathbf{t}]\in SE(3)\)
(from camera to body/base) is available, points in the body frame are
\(\mathbf{X}_B=\mathbf{R}\mathbf{X}+\mathbf{t}\mathbf{1}^\top\).

To compensate lens distortion, we optionally apply an undistortion map
\(\phi:\mathbb{R}^2\!\to\!\mathbb{R}^2\) to pixel coordinates
prior to caching, i.e., \(\tilde{\mathbf{U}}\leftarrow\tilde{\Phi}(\tilde{\mathbf{U}})\),
where \(\tilde{\Phi}\) augments \(\phi(\cdot)\) with the homogeneous row of ones.
This keeps \(\mathbf{D}\) distortion-aware without runtime overhead.

\paragraph{Uncertainty propagation.}
Assuming independent depth noise \(\sigma_d^2\) (per sampled pixel),
the covariance of each 3D point \(\mathbf{x}_i\) follows first-order propagation:
\begin{align}
\mathbf{\Sigma}_{\mathbf{x}_i}\;\approx\;
\mathbf{J}_i\,\sigma_d^2\,\mathbf{J}_i^\top,\quad
\mathbf{J}_i=\frac{\partial \mathbf{x}_i}{\partial d_i}
=\frac{1}{s_{\text{depth}}}\,\mathbf{K}^{-1}\tilde{\mathbf{u}}_i,
\end{align}
which we use to weight downstream feature aggregation (uncertainty-aware pooling).

\paragraph{Geometric feature extraction.}
From \(\mathbf{X}\) we construct a sparse point set
\(\mathcal{P}=\{\mathbf{x}_i\}_{i=1}^{N}\subset\mathbb{R}^3\).
To balance coverage and efficiency, we optionally apply
farthest-point sampling \(\Pi_{\text{FPS}}\) or voxel hashing \(h(\cdot)\),
producing \(\tilde{\mathcal{P}}=\Pi_{\text{FPS}}(\mathcal{P})\) (or \(h(\mathcal{P})\)).
Hierarchical point features are then extracted via PointNet++~\cite{qi2017pointnetplusplus}:
\begin{align}
\mathbf{F}_{\text{D}}=\operatorname{PN++}(\tilde{\mathcal{P}})
\;\in\;\mathbb{R}^{M\times d_D},
\end{align}
where \(M\le N\) is the retained set size and \(d_D\) is the depth feature dimensionality.


\begin{figure}[t] 
    \centering
    \includegraphics[width=0.35\textwidth]{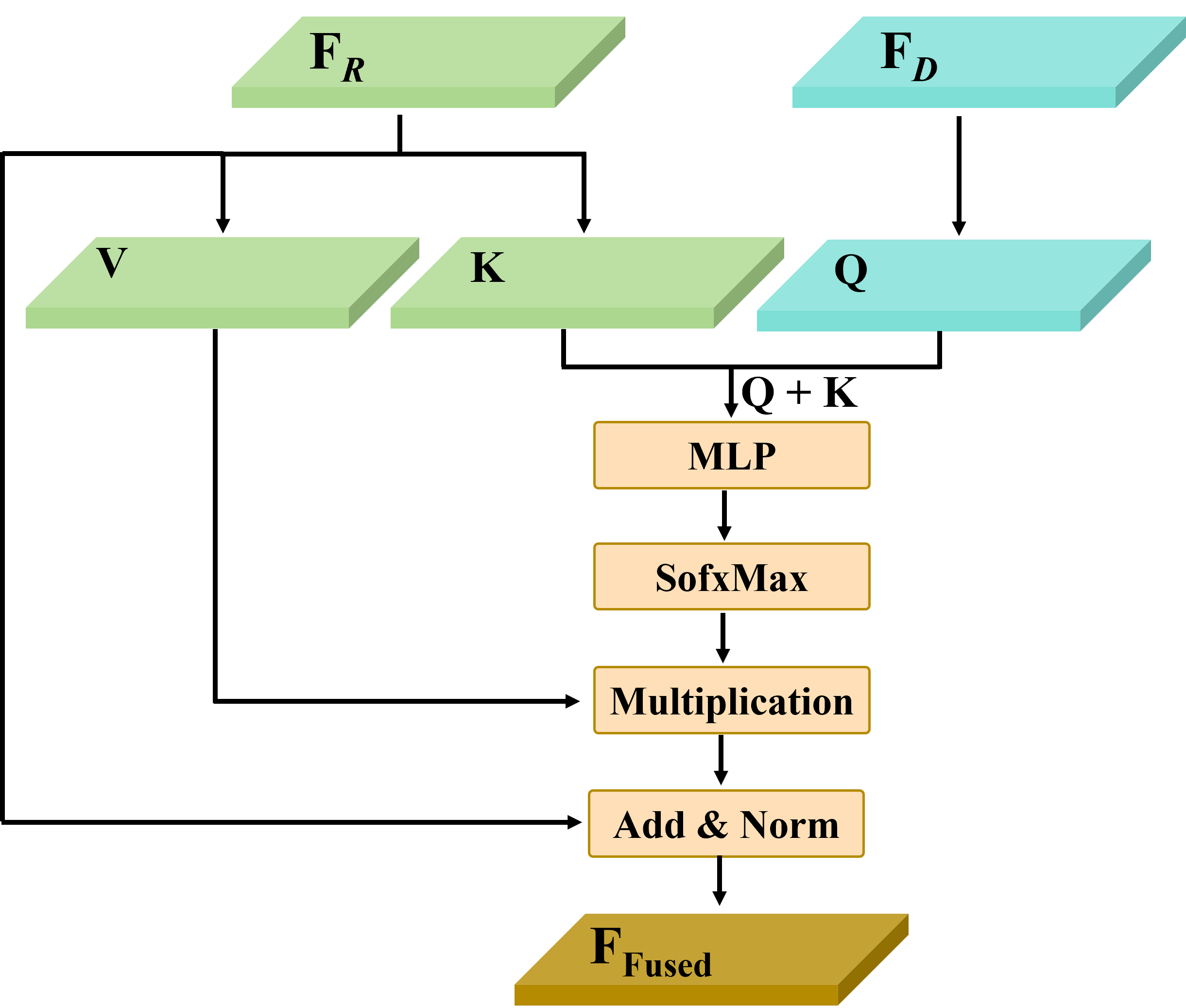} 
    \caption{Architecture of cross-attention mechanism.}
    \label{fig:sample}
\end{figure}

\textbf{Cross-Attention Mechanism:} 
Fig. 3 provides a detailed illustration of the cross-attention architecture. Let the depth-stream features be 
\(\mathbf{F}_D \in \mathbb{R}^{N \times d_D}\) 
and the RGB-stream features be 
\(\mathbf{F}_R \in \mathbb{R}^{M \times d_R}\). 
We first project them into a shared latent space of dimension \(d\) via
\begin{align}
\mathbf{Q} = \mathbf{F}_D \mathbf{W}_Q,\quad 
\mathbf{K} = \mathbf{F}_R \mathbf{W}_K,\quad
\mathbf{V} = \mathbf{F}_R \mathbf{W}_V,
\end{align}
where \(\mathbf{W}_Q,\mathbf{W}_K,\mathbf{W}_V \in \mathbb{R}^{d_{\cdot}\times d}\) are trainable matrices.  

Instead of the standard dot-product similarity, we employ an MLP-based non-linear correlation function \(\phi(\cdot,\cdot)\) between queries and keys. The attention matrix is therefore defined as
\begin{align}
\mathbf{A}_{ij} = \frac{\exp\!\left(\phi(\mathbf{Q}_i,\mathbf{K}_j)\right)}
{\sum_{j'=1}^M \exp\!\left(\phi(\mathbf{Q}_i,\mathbf{K}_{j'})\right)}, 
\quad \mathbf{A}\in\mathbb{R}^{N\times M}.
\end{align}

The fused cross-modal representation is obtained by
\begin{equation}
\label{eq:crossatt}
\mathbf{F}_{\text{Cross}} = \mathbf{A}\mathbf{V},
\end{equation}
and the final output after residual addition and layer normalization is
\begin{equation}
\label{eq:fused}
\mathbf{F}_{\text{Fused}} = \operatorname{LN}\!\left(\mathbf{F}_R + \mathbf{F}_{\text{Cross}}\right).
\end{equation}

Here, \(\phi(\cdot,\cdot)\) is implemented as a lightweight MLP that captures higher-order interactions between queries and keys, enabling richer geometric modulation than conventional linear dot-product attention. This design allows the depth-derived queries to selectively attend to semantically aligned RGB features, resulting in a fused representation \(\mathbf{F}_{\text{Fused}}\) that jointly encodes structural geometry and appearance context.

\textbf{Decoder:} 
We employ a dual-decoder architecture for autoregressive path generation. 
Let the fused cross-modal representation be 
\(\mathbf{F}_{\text{Fused}} \in \mathbb{R}^{M \times d}\), 
which is used as the shared context feature map. 
This feature map serves as the \emph{key} and \emph{value} inputs 
for both trajectory and pose decoders.

Each decoder generates a sequence of length \(T\): 
the trajectory decoder predicts position vectors 
\(\hat{\mathbf{p}} = \{\hat{\mathbf{p}}_t \in \mathbb{R}^2\}_{t=1}^T\), 
while the orientation decoder predicts unit vectors 
\(\hat{\mathbf{o}} = \{ \hat{\mathbf{o}}_t \in \mathbb{R}^2 \}_{t=1}^T\). 
The generation is autoregressive: at step \(t\), the decoder input query is
\begin{align}
\mathbf{q}_t = \begin{cases}
\mathbf{e}_{\text{start}}, & t=1, \\
\hat{\mathbf{y}}_{t-1}, & t > 1,
\end{cases}
\end{align}
where \(\mathbf{e}_{\text{start}}\) is a learnable start token 
and \(\hat{\mathbf{y}}_{t-1}\) is the decoder output from the previous step.

During training, we adopt \emph{teacher forcing}: instead of 
feeding the predicted output, we provide the ground-truth sequence 
\(\mathbf{y}_{1:T}\), augmented with positional embeddings, 
as the input queries to the decoders. This enables efficient 
parallel training while preserving the autoregressive formulation. 
To enforce temporal causality, a causal attention mask 
\(\mathbf{M}_{\text{causal}}\in\{0,-\infty\}^{T\times T}\) 
is applied in the self-attention layers, ensuring that prediction 
at step \(t\) depends only on steps \(\leq t\).


\textbf{Loss Function:} 
The network outputs a trajectory sequence 
\(\hat{\mathbf{p}} = \{\hat{\mathbf{p}}_i \in \mathbb{R}^2 \}_{i=1}^N\) 
and an orientation sequence 
\(\hat{\mathbf{o}} = \{\hat{\mathbf{o}}_i \in \mathbb{R}^2 \}_{i=1}^N\). 
The ground-truth counterparts are denoted as 
\(\mathbf{p}_i\) and \(\mathbf{o}_i\), respectively.  

To address the periodicity of the orientation angle \(\psi_i\), we do not 
directly regress \(\psi_i\). Instead, each orientation is represented 
as a unit vector 
\(\mathbf{o}_i = (\cos\psi_i, \sin\psi_i)\). 
The decoder predicts a 2D vector that is subsequently 
normalized via the L2 norm to ensure it lies on the unit circle.

The overall training objective is a weighted sum of the position and 
orientation errors:
\begin{equation}
\label{eq:loss}
\mathcal{L}_{\text{total}}
= \alpha \cdot \frac{1}{N}\sum_{i=1}^N 
\|\mathbf{p}_i - \hat{\mathbf{p}}_i\|_1
+ \beta \cdot \frac{1}{N}\sum_{i=1}^N 
\|\mathbf{o}_i - \hat{\mathbf{o}}_i\|_1,
\end{equation}
where \(\alpha\) and \(\beta\) are trade-off coefficients between position 
and orientation accuracy. After empirical tuning, we set 
\(\alpha = 0.63\) and \(\beta = 0.37\), which achieves a balanced 
performance across the two objectives.



\begin{figure}[t] 
    \centering
    \includegraphics[width=0.47\textwidth]{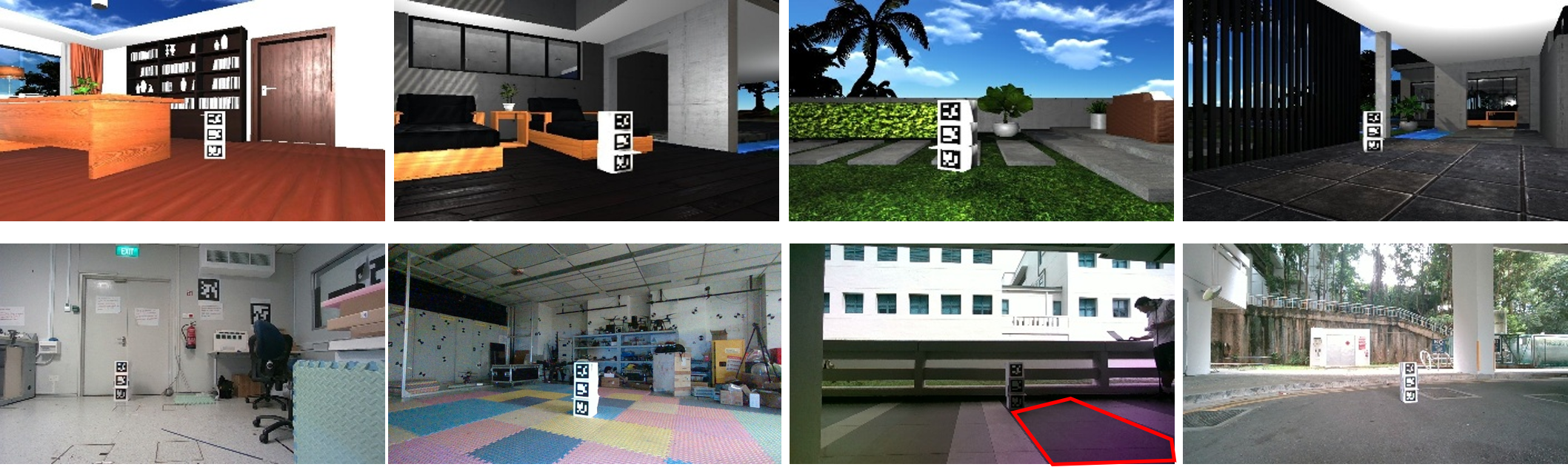} 
    \caption{The first row shows images from virtual scenes, while the second row features real-world images. Notably, the second image in the second row is a nighttime shot, and the red-boxed area in the third image appears purple due to infrared interference.}
    \label{fig:sample}
    \vspace{-1mm}
\end{figure}
\setlength{\textfloatsep}{10pt} 

\section{EXPERIMENTS}

\subsection{Dataset Preparation and Implementation Details}
Fig. 4 presents typical examples from our large-scale visual docking dataset, which combines virtual and real environments for both indoor and outdoor scenarios. To enhance its challenge and robustness, the dataset was significantly augmented through domain randomization, including diverse variations in lighting conditions, object textures, and sensor noise. Additionally, the visual docking dataset encompasses data from docking tasks initiated from various starting positions. This is crucial for training the DVDP model, ensuring it can achieve reliable docking from any initial position.

The virtual dataset is collected from indoor and outdoor scenarios constructed using Unity 3D. We utilize ROS2 to facilitate communication between simulation components, enabling seamless integration of sensor data and control commands. The dataset encompasses a variety of scene types, including bedrooms, sitting room, dining room, study, balcony, yard. When the robot begins its docking process, its camera sensors output RGB images and depth images. Subsequently, during the execution of the docking task, the robot outputs its position and orientation relative to its initial coordinate system at a fixed frequency. In virtual scenarios, docking path generation relies on rule-based algorithmic generation. The virtual scenario dataset comprises a total of 10,000 samples, with 8,000 used for training and 2,000 for evaluation. 

\begin{figure}[t] 
    \centering
    \includegraphics[width=0.47\textwidth]{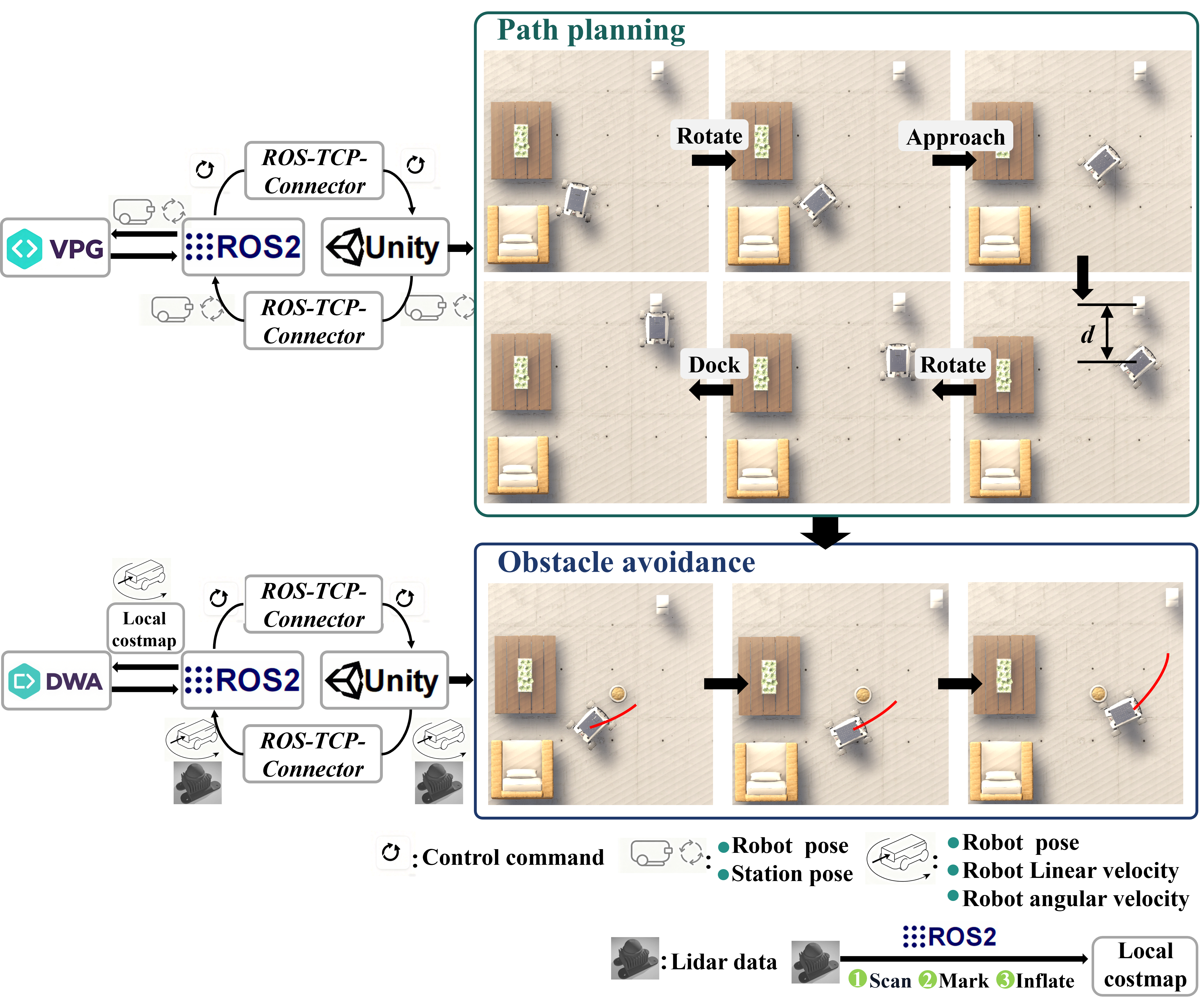} 
    \caption{Overview of our hierarchical docking architecture. The high-level VPG planner generates a four-step policy (Rotate \textrightarrow{} Approach \textrightarrow{} Rotate \textrightarrow{} Dock) based on a virtual docking point at distance \textit{d}. This policy is then executed by the low-level DWA controller, which uses real-time Lidar data for dynamic obstacle avoidance. The architecture separates strategic planning (VPG) from reactive control (DWA) and is validated in a ROS2-Unity co-simulation environment via the \texttt{ROS-TCP-Connector}.}
    \label{fig:sample}
    \vspace{-1mm}
\end{figure}

\setlength{\textfloatsep}{10pt} 

Inspired by the research conducted by \cite{Luo2005Automatic}, we developed a rule-based autonomous docking algorithm to enhance the efficiency of robotic data collection, as illustrated in Fig. 5. Our docking approach is composed of two hierarchical components: a Virtual Point Guidance (VPG) policy for high-level path generation and the Dynamic Window Approach (DWA) for low-level motion control and obstacle avoidance. The VPG policy, which generates the desired docking trajectory, is executed in two phases: initialization and motion control. In the initialization phase, the system acquires the robot's current pose and the base station's global pose to define two critical waypoints: a Real Docking Point for the final physical connection, and a Virtual Docking Point positioned at a predefined standoff distance. Subsequently, the motion control phase executes a decoupled, four-step maneuver based on these waypoints: (1) an initial rotation to face the Virtual Docking Point, (2) a linear approach to it, (3) a final rotation to align with the required docking orientation, and (4) a terminal linear translation to the Real Docking Point. This strategy effectively decouples the task into distinct approach and alignment sub-tasks. During execution, the DWA controller continuously generates safe velocity commands to follow the path prescribed by VPG while reacting to real-time data from the MID360 LiDAR to avoid potential obstacles encountered along the path.

We collected real-world data using a modified SCOUT Mini platform with docking capabilities, equipped with an Intel RealSense D455 camera and a MID 360 LiDAR. The dataset includes scenes from laboratory, corridor, and open ground floor. In real-world scenarios, obtaining precise pose coordinates of the base station relative to the robot is challenging. Consequently, the docking paths in these environments are typically collected and generated by experts. The real-world scenario dataset comprises a total of 1200 samples, with 1000 used for training and 200 for evaluation. 

Our method is implemented using the PyTorch framework and is trained with the Adam optimizer on a single NVIDIA GeForce RTX 4090 GPU. The batch size is set to 16, and a consistent hyperparameter configuration is employed across all experiments. After training on the virtual dataset, we leverage these results to train on real-world data. This approach facilitates the transition from virtual to real scenarios and effectively mitigates the issue of limited real data availability.

\begin{figure*}[htbp] 
    \centering 
    \includegraphics[width=0.93\textwidth]{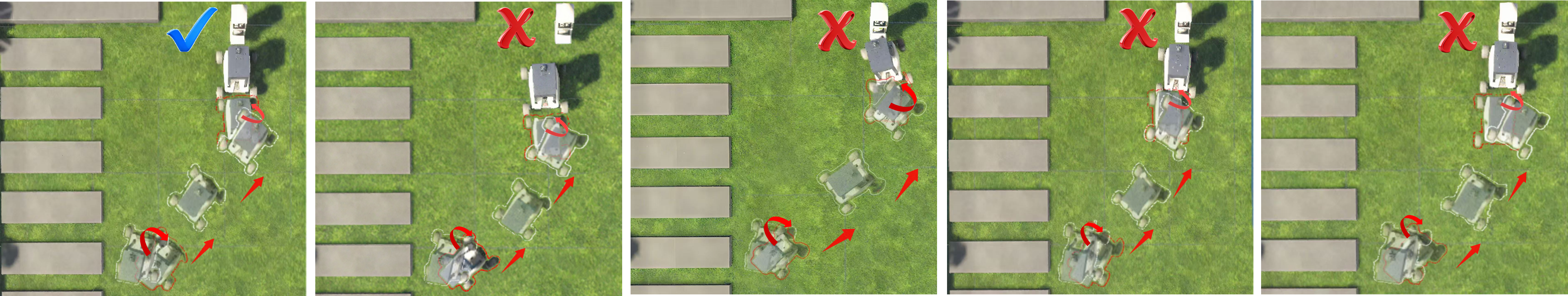} 
    \caption{Typical visualized result of the comparative experiment results. From left to right: DVDP, Centerpose \cite{9812299}, AnyGrasp \cite{9156992}, PVN3D \cite{9157414} and Hoang \textit{et al.} \cite{10430220}.}
    \label{fig:two_column_center}
\end{figure*}

\subsection{Evaluation Metric}
We have defined several metrics to evaluate the inference capability of the end-to-end visual docking model, as detailed below:

\textbf{L2 Distance (L2 Dis.)}: The L2 distance measures the mean Euclidean distance between corresponding trajectory point coordinates of the predicted and actual trajectories. This metric is employed to assess the precision and accuracy of a model's predictions for trajectory point coordinates during inference.

\textbf{Average Orientation Error (AER)}: The Average Orientation Error measures the average error in robot orientation at each trajectory point between predicted and actual trajectories. This metric is designed to assess the precision and accuracy of the model's predictions regarding robot orientation during inference. 

\textbf{Final Docking Position Error (FDPE)}: The Final Docking Position Error describes the discrepancy between the predicted and actual trajectory point coordinates when a robot completes its final docking maneuver. 

\textbf{Final Docking Orientation Error (FDOE)}: The Final Docking Orientation Error quantifies the discrepancy between the predicted and actual orientations of the robot at the final docking maneuver.   

\textbf{Success Rate (SR)}: The Success Rate measures the percentage of trials considered successful. A trial is deemed successful only if the final state meets the predefined accuracy thresholds (i.e., FDPE $<$ 0.05m and FDOE $<$ $5^\circ$) and the entire predicted trajectory is confirmed to be collision-free and kinematically plausible. As a holistic metric, SR evaluates the practical reliability of our method by integrating both the precision of the final docking pose and the safety of the generated path.

\begin{figure*}[htbp] 
    \centering 
    \includegraphics[width=0.93\textwidth]{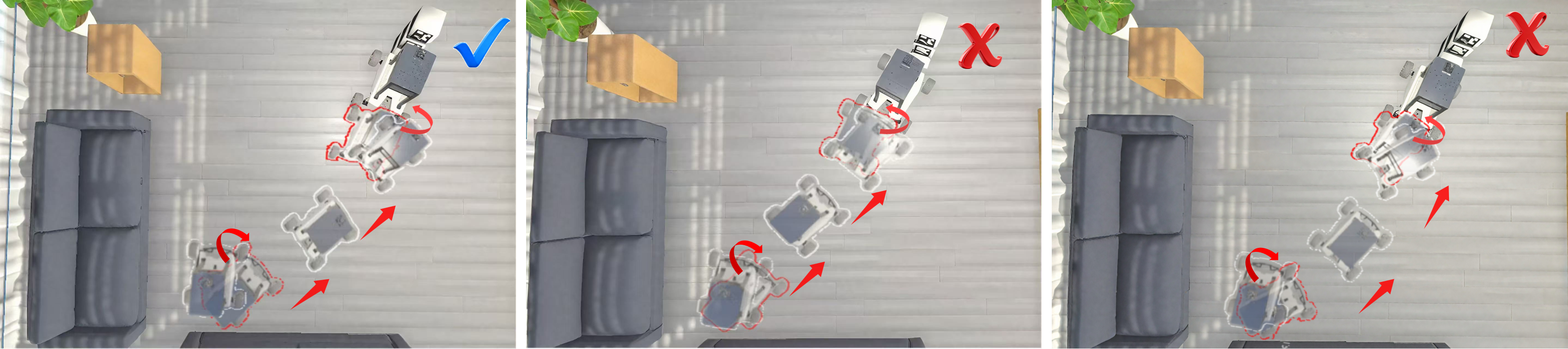} 
    \caption{Typical visualized result of the ablation experiment results. From left to right: DVDP, w/o CrossAtt and w/o Decoder.}
    \label{fig:two_column_center}
    \vspace{-2mm}
\end{figure*}

\subsection{Comparative Experiment}
For a fair and systematic comparison, we selected state-of-the-art (SOTA) methods from related domains like 6D pose estimation and grasping (e.g., Centerpose \cite{9812299}, AnyGrasp \cite{9156992}, PVN3D \cite{9157414} and Hoang \textit{et al.} \cite{10430220}) to serve as baseline encoders, chosen for their representative and diverse architectures. As these models were not originally designed for docking, we did not evaluate them directly. Instead, we re-engineered them by integrating their respective encoders into our DVDP framework, which provides a common autoregressive decoder for all variants. After retraining each hybrid model from scratch on our dataset, this methodology allows us to isolate the performance contribution of each perception backbone within a unified end-to-end trajectory generation context, ensuring all methods are compared on equal footing. This provides the first unified evaluation of these diverse perception backbones for the task of docking. 

Table I shows the results of the comparative experiment. The poor performance of CenterPose and AnyGrasp can be attributed to their unimodal design (RGB images or depth maps, respectively), which struggles to extract the rich, coupled features of pose and context required for docking. When comparing against other multi-modal methods like PVN3 and Hoang \textit{et al.}, DVDP's superior performance stems from a key architectural difference in the backbone network. While these competing methods also employ a Pyramid Pooling Module (PPM) for contextual feature capture, their ResNet-based backbones lack an intrinsic mechanism for inter-layer feature aggregation. Consequently, the features fed into their PPMs are already deficient in hierarchical information. In contrast, DVDP addresses this limitation by first leveraging a DLA backbone for effective hierarchical feature fusion, and only then feeding this deeply fused representation into a PPM for contextual enrichment. This synergistic "hierarchical-then-contextual" design enables DVDP to generate a significantly more comprehensive and discriminative feature representation—one that encapsulates both fine-grained geometric details and a global understanding of the scene. This, in turn, provides a richer basis for the downstream trajectory generation module, explaining its superior performance.

\begin{table}[t]
\centering
\caption{Comparative Experiment Results.}
\setlength{\tabcolsep}{4pt} 
\renewcommand{\arraystretch}{1.2} 
\begin{tabular}{lccccc}
\toprule
\textbf{Method} & \textbf{L2 Dis. \(\downarrow\)} & \textbf{AER \(\downarrow\)} & \textbf{FDPE \(\downarrow\)} & \textbf{FDOE \(\downarrow\)} & \textbf{SR \(\uparrow\)}\\
\midrule
\textbf{DVDP (Ours)} & \textbf{0.044m} & \textbf{4.6$^\circ$} & \textbf{0.0445m} & \textbf{4.5$^\circ$} & \textbf{73.2\%} \\
CenterPose \cite{9812299} & 0.070m & 6.2$^\circ$ & 0.073m & 5.9$^\circ$  & 27.1\% \\
AnyGrasp \cite{9156992}   & 0.067m & 7.3$^\circ$ & 0.065m & 6.7$^\circ$ & 28.9\% \\
PVN3D \cite{9157414}      & 0.055m & 5.3$^\circ$  & 0.057m & 5.6$^\circ$  & 46.5\% \\
Hoang \textit{et al.} \cite{10430220} & 0.051m & 4.9$^\circ$ & 0.047m & 5.0$^\circ$ & 53.3\% \\
\bottomrule
\end{tabular}
\label{tab:comparative_results}
\end{table}

Fig. 6 presents a typical visualized result of the comparative experiment results. Due to a lack of accurate depth features, CenterPose generates docking trajectories that halt just short of reaching the front of the base station. The docking failures observed with AnyGrasp can be attributed to the absence of precise base station orientation information normally provided by RGB input. Although the model is capable of reaching the target's proximity, it cannot achieve the required terminal alignment, resulting in a large pose error that prevents a successful dock. DVDP achieves state-of-the-art performance in the docking task, with a smooth trajectory during the docking process and high docking accuracy.

\subsection{Ablation Experiment}
To investigate the impact of different components on the DVDP architecture, an ablation study was conducted as shown in Table II. We replace the cross-attention mechanism with a simplified fusion module, where RGB features and depth features are summed and then normalized. The DVDP consistently outperforms the version without the cross-attention mechanism (w/o CrossAtt) across all evaluation metrics. This indicates that the cross-attention mechanism effectively integrates RGB and depth features, offering more accurate and enriched information about the base station and surrounding environment for the docking trajectory generation process. Meanwhile, we replaced our autoregressive encoder with a non-autoregressive encoder. The experimental results indicate that DVDP exhibits some improvements in performance metrics compared to the version using a non-autoregressive encoder (w/o Decoder). This suggests that the autoregressive encoder is better equipped to model state transitions and dynamic constraints during the trajectory generation process. Since trajectory generation is inherently a sequential decision-making process, the autoregressive approach naturally aligns with this temporal dependency. Consequently, the model is able to learn trajectories that are smoother and more consistent with physical laws.

Fig. 7 illustrates a typical example of the ablation experiment results. Compared to DVDP, the feature fusion module used in w/o CrossAtt is less effective. Its output feature maps fail to provide sufficiently accurate information for the subsequent trajectory generation module, resulting in larger deviations from the actual expert trajectories. Compared to the w/o Decoder version, DVDP demonstrates superior docking trajectory generation. The non-autoregressive approach, due to its parallel generation of all trajectory points, lacks explicit modeling of temporal dependencies between states. This can lead to discontinuities or kinematically infeasible transitions in the trajectory. In contrast, DVDP employs an autoregressive decoder that generates each point sequentially, grounding each prediction on the history of the trajectory. This inherent causal structure ensures that the generated trajectories are not only globally goal-directed but also maintain local smoothness and kinematic feasibility.

\begin{table}[t]
\centering
\caption{Ablation Study Results.}
\setlength{\tabcolsep}{4pt} 
\renewcommand{\arraystretch}{1.1} 

\begin{tabular}{lccccc}
\toprule
\textbf{Method} & \textbf{L2 Dis. \(\downarrow\)} & \textbf{AER \(\downarrow\)} & \textbf{FDPE \(\downarrow\)} & \textbf{FDOE \(\downarrow\)} & \textbf{SR \(\uparrow\)}\\
\midrule
\textbf{DVDP (Ours)} & \textbf{0.044m} & \textbf{4.6$^\circ$} & \textbf{0.045m} & \textbf{4.5$^\circ$} & \textbf{73.2\%} \\
w/o CrossAtt & 0.064m & 5.8$^\circ$ & 0.064m & 5.9$^\circ$ & 35.6\% \\
w/o Decoder  & 0.048m & 4.8$^\circ$ & 0.047m & 4.7$^\circ$ & 66.8\% \\
\bottomrule
\end{tabular}
\label{tab:ablation_results}
\end{table}

\subsection{Real Robot Experiment}
To evaluate the effectiveness of DVDP in real-world visual docking on a mobile robot, we deployed the DVDP network onto a modified SCOUT Mini mobile robotic platform. This platform is equipped with an 11th Gen Intel Core i7-1165G7 processor and operates on the Ubuntu 20.04 system. Fig. 8 illustrates the SCOUT Mini robot performing a DVDP-based visual docking task in a laboratory setting. Utilizing the NVIDIA D455 stereo camera mounted on the SCOUT Mini, RGB and depth images are captured at the initiation of the docking procedure. These images are processed by the DVDP network and subsequently handed over to the lower-level controller, which directs the robot's movement. This process successfully completes the visual docking task with the real mobile robot. Moreover, we plan to expand the mobile robot visual docking datasets, especially the real-world dataset, aiming to enhance the stability when deployed on real robots.
\begin{figure}[t] 
    \centering
    \includegraphics[width=0.47\textwidth]{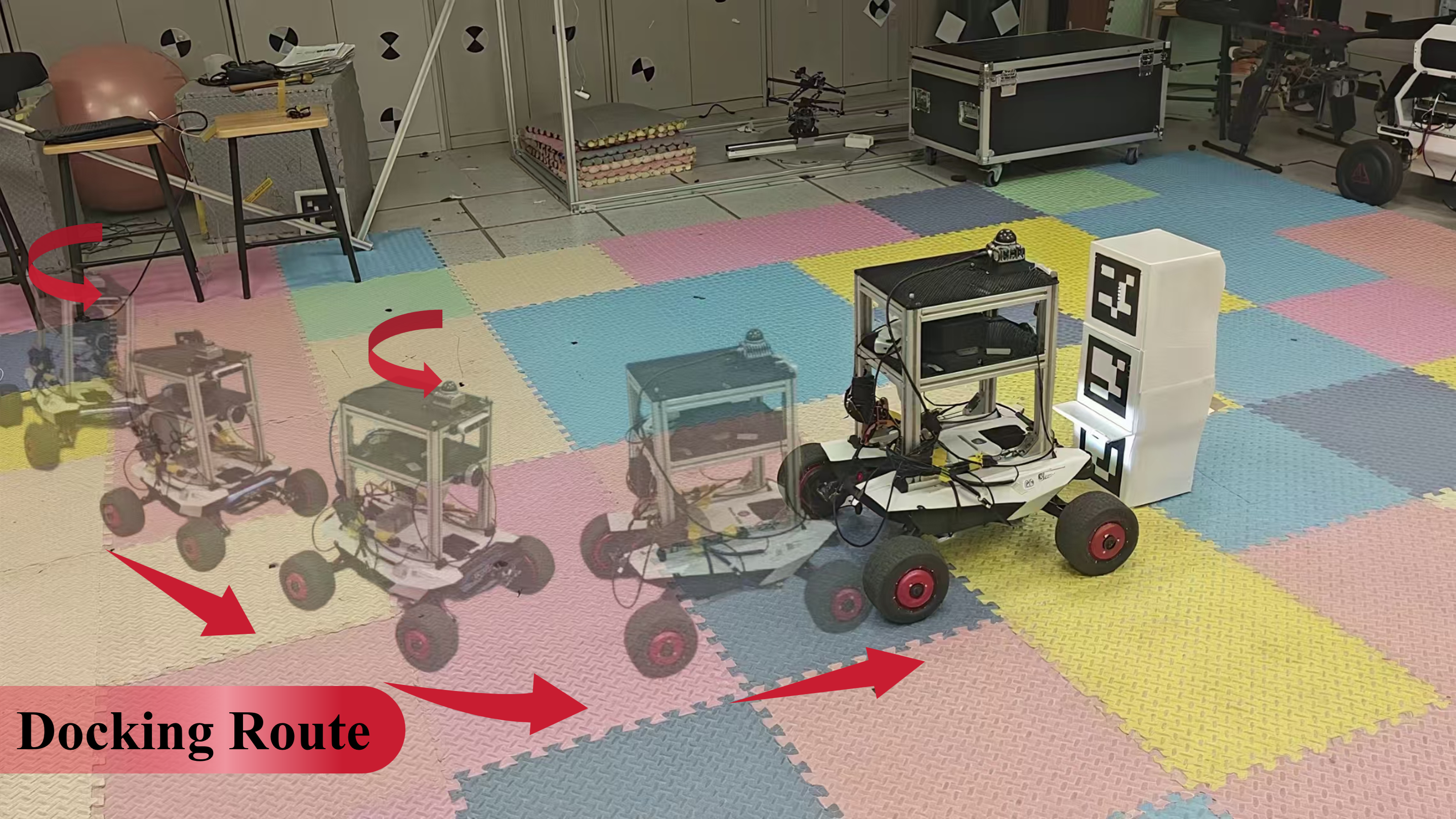} 
    \caption{Real-world visual docking experiment.}
    \label{fig:sample}
\end{figure}

\section{CONCLUSIONS}
In this paper, we highlight that existing visual docking methods often fail, as visual features are prone to distortion or occlusion. In response to this limitation, we propose DVDP, an end-to-end strategy that generates a complete docking trajectory directly from RGB and depth images to achieve docking. Additionally, we developed a hybrid real-virtual indoor and outdoor mobile robot automatic docking dataset to support research in this area. Through extensive comparative and ablation studies, we demonstrate that DVDP represents the state-of-the-art in end-to-end visual docking policies for mobile robots. In the future, we aim to expand our mobile robot automatic docking dataset to improve stability in real-world deployments. 

\section{ACKNOWLEDGMENTS}
The authors acknowledge the use of large language models (e.g., Gemini 2.5 Pro by Google DeepMind) for assisting in improving the clarity of the manuscript. All technical contributions and interpretations remain the responsibility of the authors.

\addtolength{\textheight}{0cm}   





\bibliographystyle{IEEEtran}
\bibliography{refs}

\begin{thebibliography}{10}
\providecommand{\url}[1]{#1}
\csname url@samestyle\endcsname
\providecommand{\newblock}{\relax}
\providecommand{\bibinfo}[2]{#2}
\providecommand{\BIBentrySTDinterwordspacing}{\spaceskip=0pt\relax}
\providecommand{\BIBentryALTinterwordstretchfactor}{4}
\providecommand{\BIBentryALTinterwordspacing}{\spaceskip=\fontdimen2\font plus
\BIBentryALTinterwordstretchfactor\fontdimen3\font minus \fontdimen4\font\relax}
\providecommand{\BIBforeignlanguage}[2]{{%
\expandafter\ifx\csname l@#1\endcsname\relax
\typeout{** WARNING: IEEEtran.bst: No hyphenation pattern has been}%
\typeout{** loaded for the language `#1'. Using the pattern for}%
\typeout{** the default language instead.}%
\else
\language=\csname l@#1\endcsname
\fi
#2}}
\providecommand{\BIBdecl}{\relax}
\BIBdecl

\bibitem{zhou2023autonomous}
Y.~Zhou, Y.~He, Y.~Yan, F.~Li, N.~Li, C.~Zhang, Z.~Lu, and Z.~Yang, ``Autonomous charging docking control method for unmanned vehicles based on vision and infrared,'' in \emph{J. Phys. Conf. Ser.}, vol. 2584, no.~1, 2023, p. 012065.

\bibitem{2015Calibration}
L.~Marton, C.~Nagy, Z.~Biro-Ambrus, and K.~Gyrgy, ``Calibration and measurement processing for ultrasonic indoor mobile robot localization systems,'' \emph{Proc. IEEE Int. Conf. Ind. Technol. (ICIT)}, 2015.

\bibitem{2024A}
S.~Xu, Y.~Jiang, Y.~Li, B.~Wang, T.~Xie, S.~Li, H.~Qi, A.~Li, and J.~Cao, ``A stereo visual navigation method for docking autonomous underwater vehicles,'' \emph{J. Field Robot.}, no.~2, p.~41, 2024.

\bibitem{2025Machine}
D.~Worth, J.~Choate, R.~Raettig, S.~Nykl, and C.~Taylor, ``Machine visual perception from sim-to-real transfer learning for autonomous docking maneuvers,'' \emph{Neural Comput. Appl.}, vol.~37, no.~4, 2025.

\bibitem{2024Vision}
X.~Fangyi, X.~U. Cheng, L.~Jinguo, and X.~Zhihui, ``Vision-based docking system for an aromatic-hydrocarbon-inspired reconfigurable robot,'' \emph{Sci. China Technol. Sci.}, no.~6, 2024.

\bibitem{2024Vision11}
C.~E. Mower, M.~Huber, H.~Tian, A.~Davoodi, E.~V. Poorten, T.~Vercauteren, and C.~Bergeles, ``Vision and contact based optimal control for autonomous trocar docking,'' \emph{arXiv}, 2024.

\bibitem{Zhang:22}
Y.~Zhang, X.~Wang, P.~Lei, S.~Wang, Y.~Yang, L.~Sun, and Y.~Zhou, ``Smart vector-inspired optical vision guiding method for autonomous underwater vehicle docking and formation,'' \emph{Opt. Lett.}, vol.~47, no.~11, pp. 2919--2922, 2022.

\bibitem{2022Automated}
A.~J. Choi, J.~Park, and J.~H. Han, ``Automated aerial docking system using onboard vision-based deep learning,'' \emph{J. Aerosp. Inf. Syst.}, vol.~19, no.~6, p.~16, 2022.

\bibitem{2019Autonomous}
Z.~Yan, P.~Gong, W.~Zhang, Z.~Li, and Y.~Teng, ``Autonomous underwater vehicle vision guided docking experiments based on l-shaped light array,'' \emph{IEEE Access}, 2019.

\bibitem{Volden2022Visionbased}
{\O}.~Volden, A.~Stahl, and T.~I. Fossen, ``Vision-based positioning system for auto-docking of unmanned surface vehicles (usvs),'' \emph{International Journal of Intelligent Robotics and Applications}, vol.~6, no.~1, pp. 86--103, 2022.

\bibitem{perez2021robot}
C.~P{\'e}rez-D’Arpino, C.~Liu, P.~Goebel, R.~Mart{\'\i}n-Mart{\'\i}n, and S.~Savarese, ``Robot navigation in constrained pedestrian environments using reinforcement learning,'' in \emph{Proc. IEEE Int. Conf. Robot. Autom. (ICRA)}, 2021, pp. 1140--1146.

\bibitem{kulhanek2021visual}
J.~Kulh{\'a}nek, E.~Derner, and R.~Babu{\v{s}}ka, ``Visual navigation in real-world indoor environments using end-to-end deep reinforcement learning,'' \emph{IEEE Robot. Autom. Lett.}, vol.~6, no.~3, pp. 4345--4352, 2021.

\bibitem{Kotar_2023_ICCV}
K.~Kotar, A.~Walsman, and R.~Mottaghi, ``Entl: Embodied navigation trajectory learner,'' in \emph{Proc. IEEE/CVF Int. Conf. Comput. Vis. (ICCV)}, October 2023, pp. 10\,863--10\,872.

\bibitem{liu2024volumetric}
R.~Liu, W.~Wang, and Y.~Yang, ``Volumetric environment representation for vision-language navigation,'' in \emph{Proc. IEEE/CVF Conf. Comput. Vis. Pattern Recognit. (CVPR)}, 2024, pp. 16\,317--16\,328.

\bibitem{10855557}
M.~Seo, H.~A. Park, S.~Yuan, Y.~Zhu, and L.~Sentis, ``Legato: Cross-embodiment imitation using a grasping tool,'' \emph{IEEE Robot. Autom. Lett.}, vol.~10, no.~3, pp. 2854--2861, 2025.

\bibitem{10854684}
R.~Freiberg, A.~Qualmann, N.~A. Vien, and G.~Neumann, ``Diffusion for multi-embodiment grasping,'' \emph{IEEE Robot. Autom. Lett.}, vol.~10, no.~3, pp. 2694--2701, 2025.

\bibitem{liu2024efficient}
Y.~Liu, A.~Qualmann, Z.~Yu, M.~Gabriel, P.~Schillinger, M.~Spies, N.~A. Vien, and A.~Geiger, ``Efficient end-to-end detection of 6-dof grasps for robotic bin picking,'' in \emph{Proc. IEEE Int. Conf. Robot. Autom. (ICRA)}, 2024, pp. 5427--5433.

\bibitem{Chen2024FusedNet}
F.~Chen, Y.~Tang, C.~Tai \emph{et~al.}, ``Fusednet: End-to-end mobile robot relocalization in dynamic large-scale scene,'' \emph{IEEE Robot. Autom. Lett.}, vol.~9, no.~5, pp. 4099--4105, 2024.

\bibitem{PAL2024104567}
A.~Pal, A.~C. Leite, and P.~J. From, ``A novel end-to-end vision-based architecture for agricultural human–robot collaboration in fruit picking operations,'' \emph{Robot. Auton. Syst.}, vol. 172, p. 104567, 2024.

\bibitem{10610200}
X.~Cheng, K.~Shi, A.~Agarwal, and D.~Pathak, ``Extreme parkour with legged robots,'' in \emph{Proc. IEEE Int. Conf. Robot. Autom. (ICRA)}, 2024, pp. 11\,443--11\,450.

\bibitem{yu2018deep}
F.~Yu, D.~Wang, E.~Shelhamer, and T.~Darrell, ``Deep layer aggregation,'' in \emph{Proc. IEEE/CVF Conf. Comput. Vis. Pattern Recognit. (CVPR)}, 2018, pp. 2403--2412.

\bibitem{9812299}
Y.~Lin, J.~Tremblay, S.~Tyree, P.~A. Vela, and S.~Birchfield, ``Single-stage keypoint- based category-level object pose estimation from an rgb image,'' in \emph{Proc. IEEE Int. Conf. Robot. Autom. (ICRA)}, 2022, pp. 1547--1553.

\bibitem{zhao2017pyramid}
H.~Zhao, J.~Shi, X.~Qi, X.~Wang, and J.~Jia, ``Pyramid scene parsing network,'' in \emph{Proc. IEEE/CVF Conf. Comput. Vis. Pattern Recognit. (CVPR)}, 2017, pp. 2881--2890.

\bibitem{qi2017pointnetplusplus}
C.~R. Qi, L.~Yi, H.~Su, and L.~J. Guibas, ``Pointnet++: Deep hierarchical feature learning on point sets in a metric space,'' in \emph{Proc. Adv. Neural Inf. Process. Syst. (NeurIPS)}, vol.~30, 2017.

\bibitem{Luo2005Automatic}
R.~C. Luo, C.~T. Liao, K.~L. Su, and K.~C. Lin, ``Automatic docking and recharging system for autonomous security robot,'' in \emph{Proc. IEEE/RSJ Int. Conf. Intell. Robot. Syst. (IROS)}.\hskip 1em plus 0.5em minus 0.4em\relax IEEE, 2005.

\bibitem{9156992}
H.~S. Fang, C.~Wang, and H.~F. G. L. Y. L.~X. Lu, ``Anygrasp: Robust and efficient grasp perception in spatial and temporal domains,'' \emph{IEEE Trans. Robot.}, vol.~39, no.~5, pp. 3929--3945, 2023.

\bibitem{9157414}
Y.~He, W.~Sun, H.~Huang, J.~Liu, H.~Fan, and J.~Sun, ``Pvn3d: A deep point-wise 3d keypoints voting network for 6dof pose estimation,'' in \emph{Proc. IEEE/CVF Conf. Comput. Vis. Pattern Recognit. (CVPR)}, 2020, pp. 11\,629--11\,638.

\bibitem{10430220}
D.-C. Hoang, A.-N. Nguyen, V.-D. Vu, T.-U. Nguyen, D.-Q. Vu, P.-Q. Ngo, N.-A. Hoang, K.-T. Phan, D.-T. Tran, V.-T. Nguyen, Q.-T. Duong, N.-T. Ho, C.-T. Tran, V.-H. Duong, and A.-T. Mai, ``Graspability-aware object pose estimation in cluttered scenes,'' \emph{IEEE Robot. Autom. Lett.}, vol.~9, no.~4, pp. 3124--3130, 2024.

\end{thebibliography}

\end{document}